%% file: conference_101719.tex
\documentclass[conference]{IEEEtran}
\IEEEoverridecommandlockouts
\usepackage{cite}
\usepackage{amsmath,amssymb,amsfonts}
\usepackage{algorithmic}
\usepackage{graphicx}
\usepackage{textcomp}
\usepackage{xcolor}
\def\BibTeX{{\rm B\kern-.05em{\sc i\kern-.025em b}\kern-.08em
    T\kern-.1667em\lower.7ex\hbox{E}\kern-.125emX}}
\begin{document}

\title{Adapting Language Models to Indonesian Local Languages: An Empirical Study of Language Transferability on Zero-Shot Settings}


\author{\IEEEauthorblockN{Rifki Afina Putri}
\IEEEauthorblockA{\textit{Department of Computer Science and Electronics} \\
\textit{Universitas Gadjah Mada}\\
Yogyakarta, Indonesia \\
rifki.putri@ugm.ac.id}
}




\maketitle

\begin{abstract}
In this paper, we investigate the transferability of pre-trained language models to low-resource Indonesian local languages through the task of sentiment analysis. We evaluate both zero-shot performance and adapter-based transfer on ten local languages using models of different types: a monolingual Indonesian BERT, multilingual models such as mBERT and XLM-R, and a modular adapter-based approach called MAD-X. To better understand model behavior, we group the target languages into three categories: seen (included during pre-training), partially seen (not included but linguistically related to seen languages), and unseen (absent and unrelated in pre-training data). Our results reveal clear performance disparities across these groups: multilingual models perform best on seen languages, moderately on partially seen ones, and poorly on unseen languages. We find that MAD-X significantly improves performance, especially for seen and partially seen languages, without requiring labeled data in the target language. Additionally, we conduct a further analysis on tokenization and show that while subword fragmentation and vocabulary overlap with Indonesian correlate weakly with prediction quality, they do not fully explain the observed performance. Instead, the most consistent predictor of transfer success is the model's prior exposure to the language, either directly or through a related language.
\end{abstract}


\section{Introduction}
Language models have achieved remarkable success in multilingual NLP, enabling cross-lingual transfer learning for languages with little or no annotated data. This is particularly important for a linguistically diverse country like Indonesia, which has hundreds of local languages \cite{aji2022one}. Many of these local languages are low-resource, lacking large corpora or annotated datasets, which presents a challenge for developing language technology. A practical solution is to leverage a high-resource language (such as Indonesian or Bahasa Indonesia, the national language) or multilingual models to transfer knowledge to local languages. However, the effectiveness of such transfer can vary greatly depending on a language’s presence (or absence) in the model’s pre-training data and its linguistic similarity to languages that the model knows.

In this work, we analyze the cross-lingual transferability of language models on ten Indonesian local languages in the context of sentiment analysis using NusaX dataset \cite{winata2023nusax}. We categorize the target languages into three groups: seen languages (present in the model’s pre-training data), partially seen languages (not present in pre-training, but related to a seen language), and unseen languages (absent in pre-training with no close relatives among the training languages). This categorization allows us to study how prior exposure and linguistic relatedness affect a model's zero-shot performance. For example, Indonesian (\texttt{ind}) and Javanese (\texttt{jav}) are considered seen languages for XLM-R, since they were included in its pre-training corpus. Minangkabau (\texttt{min}), Sundanese (\texttt{sun}), and others are considered partially seen, while they were not explicitly included in pre-training, they are closely related to or have substantial lexical overlap with Indonesian or Malay. On the other hand, a language like Toba Batak (\texttt{bbc}) is unseen, as it was absent from pre-training and is not closely related to any pre-trained language.

Our core contribution is an empirical study comparing different strategies for cross-lingual transfer to these local languages. We fine-tune models on Indonesian data as the source language, and evaluate on each target language in zero-shot settings, using (1) a monolingual Indonesian BERT model (IndoBERT), (2) two multilingual models (mBERT and XLM-R), and (3) an adapter-based approach (MAD-X) that incorporates language adapters trained on unlabeled corpora. By examining the NusaX-sentiment data, we shed light on how well current language models handle low-resource languages. We also perform a tokenization analysis to see if issues like over-tokenization (breaking words into many subword pieces) correlate with drops in accuracy. Our findings confirm that zero-shot transfer is highly effective when the target language was seen during pre-training, moderately effective when only related languages were seen, and poor when the language is entirely unseen. For instance, XLM-R transfers very well to Indonesian and Javanese, and to a lesser extent to Minangkabau and Sundanese, but struggles on Toba Batak. The adapter-based MAD-X approach significantly boosts performance on some of the languages, even surpassing the full fine-tuning approach.

\section{Related Work}
\subsection{Multilingual Language Models}
Multilingual language models like mBERT \cite{devlin2019bert} and XLM-R \cite{conneau2020unsupervised} have enabled a single model to be pre-trained on dozens of languages and then fine-tuned for specific tasks. mBERT was trained on Wikipedia in 104 languages using a shared WordPiece vocabulary, and has demonstrated surprising zero-shot cross-lingual abilities using English as the language source. XLM-R improved over mBERT by training on a much larger CommonCrawl corpus (CC-100) covering 100 languages, thereby including more low-resource languages and significantly increasing data for them.

In the context of Indonesia, both mBERT and XLM-R include Indonesian, Javanese, and Sundanese in their pretraining corpora, though the latter two are represented in smaller volumes. mBERT further includes Minangkabau, although its performance is generally lower compared to XLM-R for most Indonesian local languages. In contrast, IndoBERT \cite{willie2020indonlu} is a monolingual model trained exclusively on Indonesian corpora. Although IndoBERT performs competitively on Indonesian tasks, its cross-lingual capabilities are limited by its narrower vocabulary and lack of exposure to other languages.

\subsection{Cross-Lingual Adaptation Techniques}
A major research direction for improving cross-lingual generalization focuses on adapting pretrained models to target languages. One strategy is continual pretraining, where a model is further trained on monolingual corpora of a new language using the same masked language modeling (MLM) objective. This method has been shown to improve downstream performance in both domain and language transfer scenarios \cite{gururangan2020don,liu2021continual,wang2022expanding}. Alternative approaches adapt components of the model rather than retraining the entire network. These include modifying tokenizers for morphologically rich languages \cite{kinyabert} or adapting embedding layers for new scripts using matrix factorization \cite{pfeiffer2021unk}.

In contrast to full model retraining, adapter-based approaches offer a more parameter-efficient alternative by inserting small, trainable modules into a frozen model. MAD-X \cite{pfeiffer2020madx} introduces language and task adapters that can be flexibly combined at inference time, enabling modular and scalable transfer. MAD-G \cite{ansell2021madg} extends this concept by generating multilingual adapters, reducing the need to train separate adapters for each language. Other studies show that while continual pretraining generally outperforms adapters in smaller models, adapter-based methods scale more effectively in larger architectures \cite{yong2023bloom}. Our study builds on this foundation by evaluating the zero-shot performance of multilingual and adapter-based models on underrepresented Indonesian languages and examining how language exposure and tokenization behavior influence model performance.

\section{Methodology}
\subsection{Language Categories}
We categorize the target languages based on their presence in the pre-training data of XLM-R and their linguistic proximity to seen languages. This categorization guides our analysis of cross-lingual transfer performance.

\vspace{3pt}
\noindent
\textbf{\textit{Seen Languages.}} \hspace{3pt}
Languages that the model has directly encountered during pre-training. In our case, for XLM-R these include \texttt{ind} and \texttt{jav}, since both had a presence in the CommonCrawl corpus used to train XLM-R. A language in this category is expected to have the highest transfer performance because the model has already learned its vocabulary and patterns to some extent.

\vspace{3pt}
\noindent
\textbf{\textit{Partially Seen Languages.}} \hspace{3pt}
Languages that were not part of the model's pre-training, but are closely related to a language that was. These languages often share significant vocabulary or linguistic features with a seen language. For example, Minangkabau is closely related to Malay or Indonesian, and Sundanese has influenced Indonesian vocabulary. Although the model hasn't seen these languages explicitly, it might still perform reasonably well due to similar words or structures from the seen language. We include Acehnese, Balinese, Banjarese, Buginese, Madurese, Minangkabau, Ngaju, and Sundanese in this category.

\vspace{3pt}
\noindent
\textbf{\textit{Unseen Languages.}} \hspace{3pt}
Languages that the model never saw in pre-training and that have no close linguistic relative among the pre-training languages. In our study, Toba Batak (\texttt{bbc}) is an example of an unseen language for XLM-R. We expect this category to be the most challenging for zero-shot transfer.

\begin{figure}[tbp]
\centerline{\includegraphics[width=0.95\columnwidth]{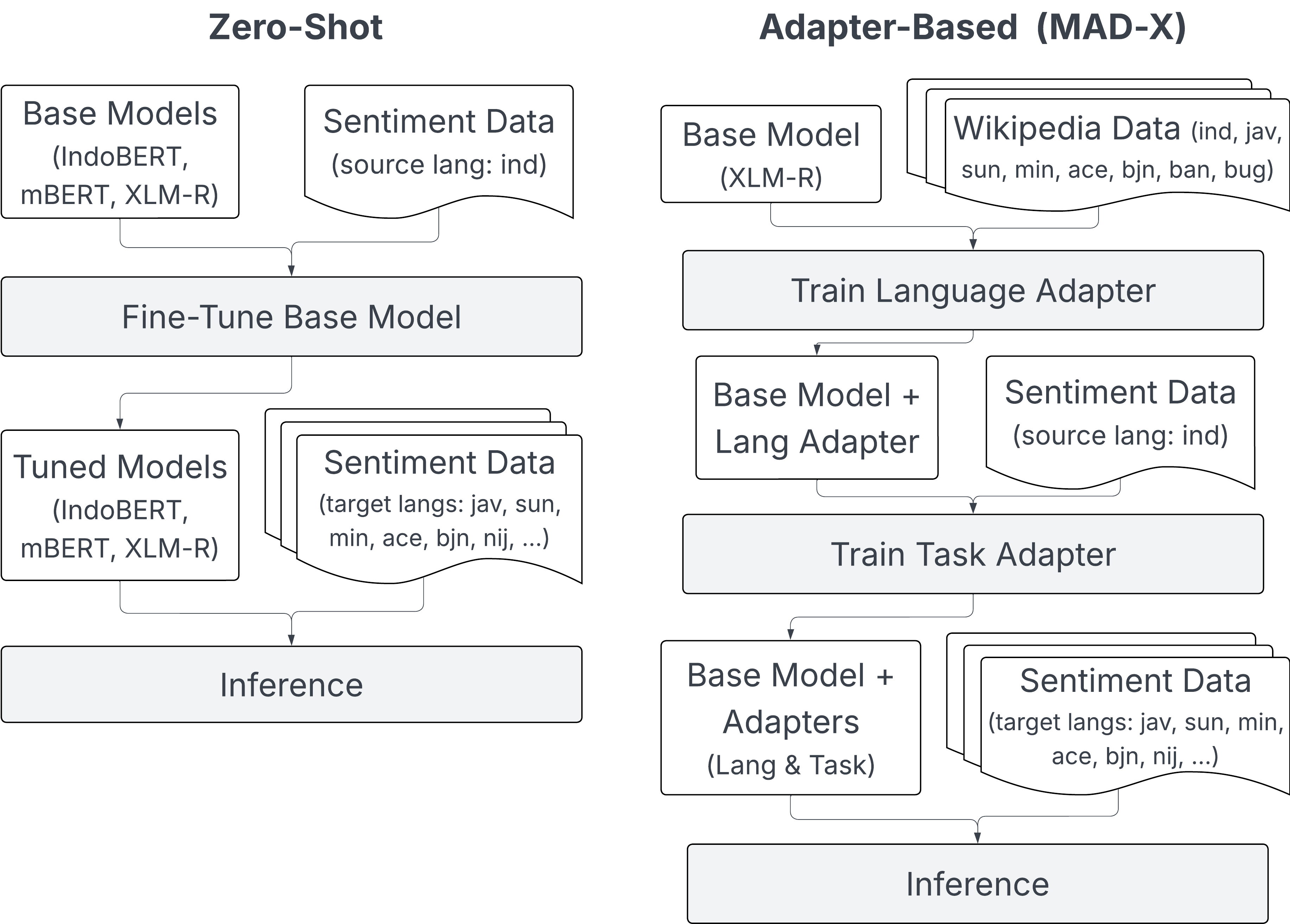}}
\caption{Illustration of zero-shot and adapter-based transfer settings.}
\label{fig:transfer_methods}
\end{figure}

\subsection{Transfer Settings}
Based on the language categorization above, we explore two transfer settings as illustrated in Figure~\ref{fig:transfer_methods}.

\vspace{3pt}
\noindent
\textbf{\textit{Zero-Shot Cross-Lingual Transfer.}} \hspace{3pt}
In this setting, we fine-tune the model on Indonesian labeled data as a source language and directly evaluate it on each target local language’s test set without any further training on the target language. This setup mimics a real-world scenario where we have annotations in a high-resource language but none in the low-resource language, relying on the model's cross-lingual ability to generalize. We apply this zero-shot setting to IndoBERT, mBERT, and XLM-R.

\vspace{3pt}
\noindent
\textbf{\textit{Adapter-Based Transfer.}} \hspace{3pt}
To further enhance cross-lingual transfer, we implement MAD-X, an adapter-based framework. First, for each target language, we train a language adapter using unlabeled text of the target language taken from Wikipedia. This is done by continuing the masked language model training objective on the target language with only the adapter parameters being updated. Separately, we fine-tune a task adapter for sentiment analysis on Indonesian data, while also using an Indonesian language adapter during this task training. At inference time, we replace the Indonesian adapter with the target language’s adapter, so that the model processes input in the target language. In essence, the task adapter learns the sentiment analysis task and the language adapter provides the language-specific vocabulary support. This approach has the benefit of utilizing unlabeled data from the target language to teach the model about that language without any labeled target examples. We implement MAD-X on top of XLM-R for all target languages where we have sufficient unlabeled text to train an adapter. The zero-shot evaluation is then done by plugging in each target's adapter.

\section{Experimental Setup}
\subsection{Dataset}
We use NusaX \cite{winata2023nusax}, a multilingual sentiment analysis dataset which contains parallel text in 12 languages: English (\texttt{eng}), Indonesian (\texttt{ind}), and 10 Indonesian local languages, namely Acehnese (\texttt{ace}), Balinese (\texttt{ban}), Banjarese (\texttt{bjn}), Buginese (\texttt{bug}), Madurese (\texttt{mad}), Minangkabau (\texttt{min}), Javanese (\texttt{jav}), Ngaju (\texttt{nij}), Sundanese (\texttt{sun}), and Toba Batak (\texttt{bbc}). We use the provided train, development, and test splits: the Indonesian portion serves as our training data for fine-tuning, and we evaluate on the test sets of each local language. We ensure no target language data is seen during training for zero-shot experiments.

Each language in the dataset has a balanced number of samples: 500 sentences for training, 100 for development, and 400 for testing. Since the dataset is parallel and balanced across all languages, any observed performance differences between languages cannot be attributed to data imbalance.

\subsection{Models}
We compare four pre-trained base model configurations:
\begin{itemize}
    \item IndoBERT \cite{willie2020indonlu}: A monolingual Indonesian BERT model. It was pre-trained on Indonesian text only (roughly 4 billion Indonesian words from news, web, and Wikipedia).
    \item mBERT \cite{devlin2019bert}: A multilingual BERT trained on Wikipedia in 104 languages. Indonesian, Javanese, and Sundanese are included among its training languages. 
    \item XLM-R \cite{conneau2020unsupervised}: A multilingual model trained on CommonCrawl data for 100 languages, including Indonesian, Javanese, Sundanese, and Minangkabau. 
    \item XLM-R with MAD-X Adapters \cite{pfeiffer2020madx}: We incorporate the MAD-X adapter approach on XLM-R as the base model. XLM-R is selected due to its better language coverage of local languages. We train a language adapter for each of the local languages that have sufficient unlabeled text available from Wikipedia, such as Indonesian (for verification), Javanese, Sundanese, Minangkabau, Acehnese, Banjarese, Balinese, and Buginese.
\end{itemize}

As a maximum performance baseline, we also evaluate full fine-tuning of XLM-R on the training set of each target local language. This represents the upper bound in terms of parameter updates and training complexity. Comparing it to the adapter-based and zero-shot configurations helps us assess the trade-offs between full fine-tuning and lightweight, parameter-efficient alternatives.

\input{tables/main_result}

\subsection{Training \& Evaluation Details}
We fine-tune all classification models using a batch size of 32 for 3 epochs, which we found sufficient for convergence on the validation set. To account for training variability, each fine-tuning experiment is repeated five times using different random seeds (0, 13, 42, 56, 99), and we report the average F1 score across runs.

For adapter-based experiments, each language adapter is trained using the masked language modeling (MLM) objective for a fixed number of steps equivalent to one epoch over the available monolingual data. Due to variation in corpus size across languages, we cap the number of training steps and cycle through smaller corpora as needed to ensure stable updates. We use the Adam optimizer with a learning rate of 5e-5 for classification fine-tuning and 1e-4 for adapter training.

All models are evaluated on the corresponding test set of each target language, and F1 score is used as the primary evaluation metric to measure performance across classes.

\begin{figure*}[htbp]
\centerline{\includegraphics[width=\textwidth]{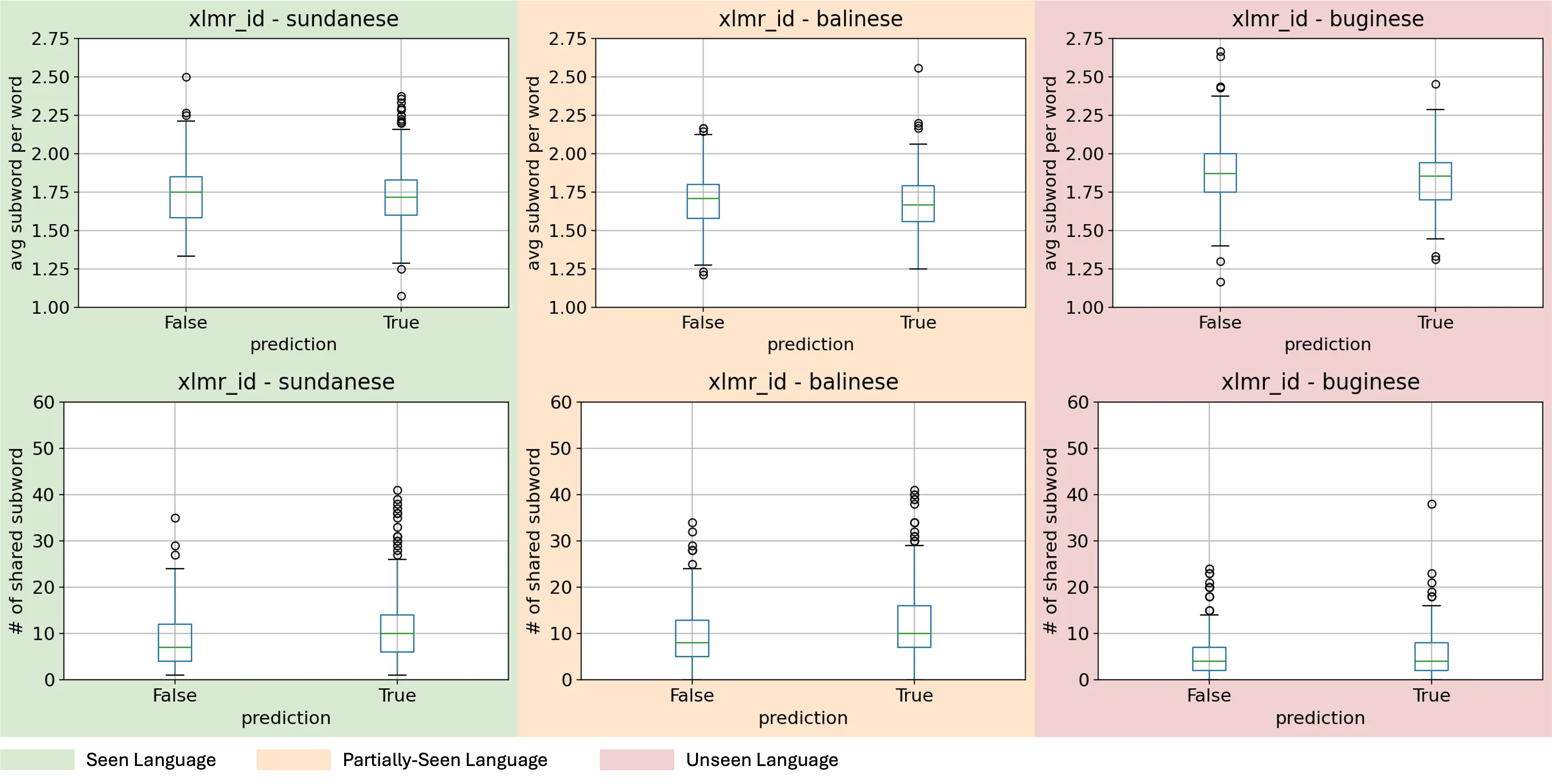}}
\caption{Tokenization analysis across three languages: Sundanese (seen), Balinese (partially seen), Buginese (unseen) using XLM-R. \textbf{Top:} Average subword per word (over-tokenization) grouped by prediction correctness. \textbf{Bottom:} Number of shared subwords with Indonesian grouped by prediction correctness.}
\label{fig:token_analysis}
\end{figure*}

\section{Results and Discussion}
\subsection{Zero-Shot Transfer Performance}
\noindent
\textbf{\textit{Perfomance on Seen vs. Unseen Languages.}} \hspace{3pt}
From Table~\ref{tab:main_result}, the zero-shot sentiment classification results clearly reflect the language category distinctions. For seen languages (Indonesian and Javanese), all models achieve strong performance, with XLM-R in particular yields the highest score. mBERT also performs reasonably well, though typically a few points lower than XLM-R, likely due to its smaller training corpus and vocabulary. IndoBERT, being monolingual, unsurprisingly performs excellently on Indonesian, but the score on other languages. This highlights the limitation of monolingual models in zero-shot transfer: without the target languages in its vocabulary or pretraining, IndoBERT cannot bridge the gap, whereas XLM-R’s multilingual training has some Javanese capability. For partially seen languages, XLM-R still significantly outperforms other models, but the gap between seen and partially seen is noticeable, except for languages with higher similarity with Indonesian such as Minangkabau and Banjarese. mBERT’s performance similarly drops for the partially seen languages, and IndoBERT is generally very poor, except on Minangkabau and Banjarese which is extremely close to Indonesian in vocabulary where IndoBERT may pick up some signals due to shared words. Unseen language performance is the lowest. In particular, for Toba Batak, XLM-R’s zero-shot accuracy is the worst among all languages. mBERT also struggles similarly. This confirms that when a language has no presence in pretraining and is not closely related to those that do, the model has a hard time transferring even the concept of sentiment labels.

\vspace{6pt}
\noindent
\textbf{\textit{MAD-X Adapter Performance.}} \hspace{3pt}
Table~\ref{tab:main_result} also shows the performance of MAD-X Adapter. The adapter-based approach yields substantial improvements on the target languages for which we trained adapters, especially those in the seen and partially seen categories. With MAD-X, we observe that XLM-R equipped with a target language adapter and the Indonesian task adapter significantly closes the gap, even surpassing the full fine-tuning baseline for Sundanese, Minangkabau, and Banjarese. The performance also improving compare to ``vanilla" zero-shot approach. For example, for Acehnese and Banjarese (partially seen languages not explicitly in pretraining), the zero-shot performance with XLM-R was moderate (F1 score of ~60-70), but with an adapter, it jumps closer to the seen language performance, into the 75-85 range. We emphasize that these adapters were trained with no sentiment labels in the target language, which means that the improvement comes purely from better language representations. In contrast, for seen languages like Indonesian and Javanese, using adapters does not significantly boost performance, likely because XLM-R already natively handles these languages well and adding an extra layer (adapter) might introduce a tiny overhead. It is still worth to note that for Sundanese, even though it also has seen during pre-training, using an adapter still improves the performance. We conjecture that this happens due to the significant number of unlabeled data in Javanese compared to Sundanese. Interestingly, for the unseen language Buginese, using an adapter actually results in lower performance. This may be attributed to the extremely limited size of its Wikipedia corpus (Buginese has only 1.8 MB, compared to 50.5 MB for Javanese and 25.1 MB for Sundanese). This suggests that adapters are most useful when the base model lacks knowledge of the target language; they serve to inject that knowledge. However, their effectiveness is also dependent on the availability and quality of unlabeled data used during adapter training.

\subsection{Subword Tokenization Analysis}
In this section, we analyze the influence of tokenization on cross-lingual sentiment classification performance. Specifically, we examine whether subword segmentation characteristics, such as over-tokenization and subword sharing with Indonesian, can explain differences in model performance across languages.

\vspace{6pt}
\noindent
\textbf{\textit{Does over-tokenization make the model give wrong predictions?}} \hspace{3pt}
We first evaluated the impact of subword fragmentation by computing the average number of subword tokens per word for each target language using the XLM-R tokenizer. The hypothesis is that languages with higher fragmentation would be harder for the model to interpret, potentially leading to lower classification performance. Our previous result shows that unseen languages like Toba Batak have lower performance, and it exhibit a high degree of token fragmentation. Conversely, partially seen languages such as Minangkabau and Banjarese have relatively higher performance and lower average subword-per-word counts. However, as shown in Fig.~\ref{fig:token_analysis}, no consistent correlation is observed at the sentence level: several misclassified sentences had low fragmentation, and others with high fragmentation were correctly classified. This suggests that while over-tokenization may contribute to reduced model understanding, it is not a decisive factor in prediction outcomes.

\vspace{3pt}
\noindent
\textbf{\textit{Does the number of shared subwords (w.r.t Indonesian) correlate with the model's predictions?}} \hspace{3pt}
To further assess the effect of lexical overlap, we computed the proportion of subword tokens in each target-language sentence that are also present in the corresponding Indonesian sentence. The rationale is that higher lexical overlap might help the model generalize from Indonesian to the target language during zero-shot inference. Interestingly, our results show no strong correlation at the sample level between shared subword proportion and prediction accuracy. As illustrated in Fig.~\ref{fig:token_analysis}, certain sentences with high subword overlap are still misclassified, while some with lower overlap are predicted correctly. This suggests that vocabulary overlap alone does not guarantee successful transfer and highlights the importance of contextual representation.

\begin{figure}[tbp]
\centerline{\includegraphics[width=0.7\columnwidth]{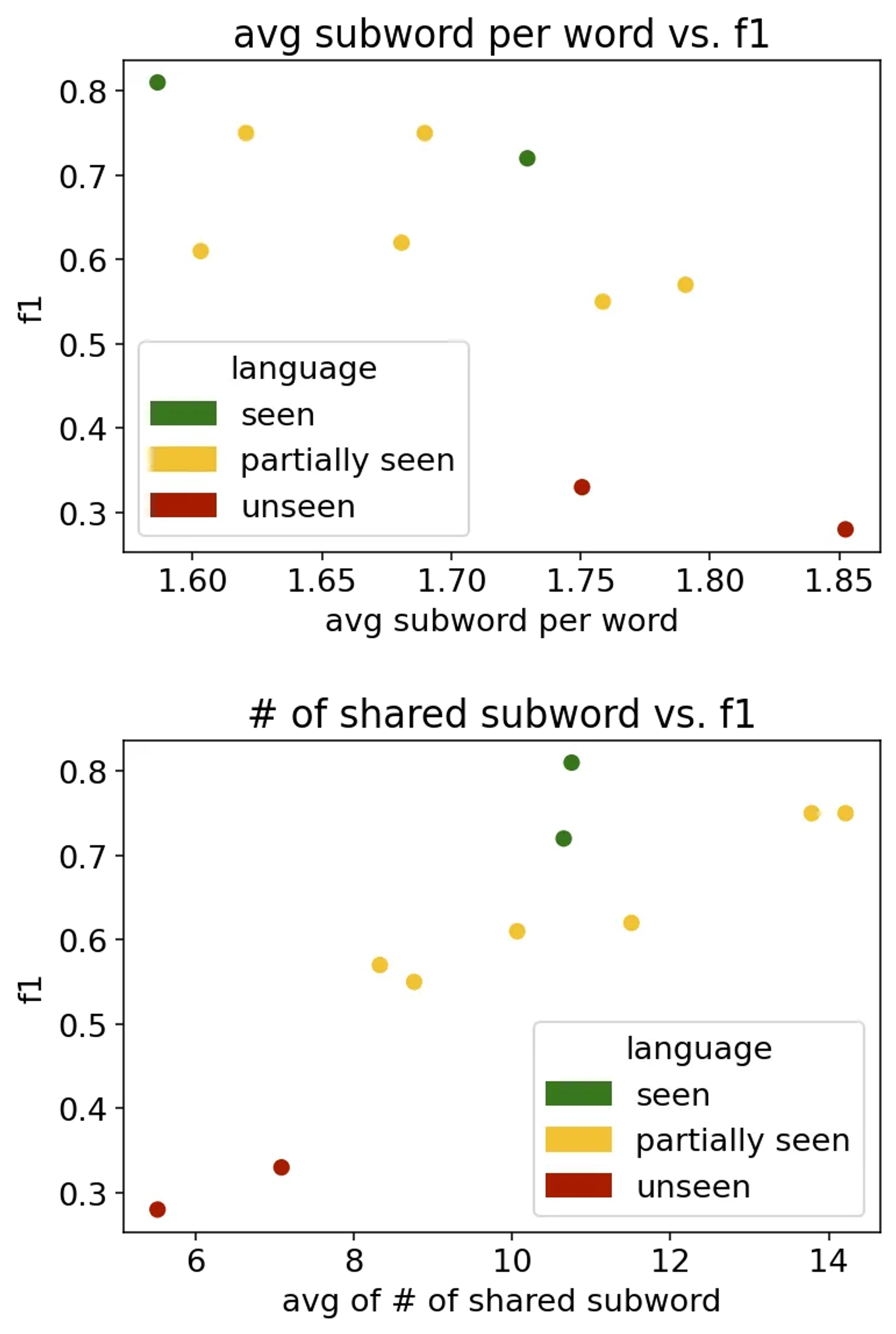}}
\caption{Language-level tokenization characteristics vs. model performance. 
\textbf{Top:} Correlation between over-tokenization (average subword per word) and F1 score. 
\textbf{Bottom:} Correlation between the number of shared subwords with Indonesian and F1 score. 
Languages are color-coded by category: seen (green), partially seen (yellow), and unseen (red).}
\label{fig:token_analysis_per_lang}
\end{figure}

\vspace{3pt}
\noindent
\textbf{\textit{How about the correlation on language-level?}} \hspace{3pt}
We aggregated both over-tokenization and shared subword statistics across all test samples for each target language and examined whether either feature correlates with language-level performance. Specifically, we analyzed whether languages with higher average subword-per-word counts or lower lexical overlap with Indonesian tend to exhibit lower classification performance.
Fig.~\ref{fig:token_analysis_per_lang} (top) shows the correlation between over-tokenization (average subwords per word) and performance, while Fig.~\ref{fig:token_analysis_per_lang} (bottom) displays the correlation between subword overlap with Indonesian and performance. A mild trend is observed in both cases: partially seen languages like Minangkabau and Banjarese, which have lower token fragmentation and higher overlap, tend to achieve better performance. Conversely, unseen languages like Buginese and Toba Batak exhibit high token fragmentation and low subword overlap, corresponding with lower F1 scores.
However, the correlations are not strong or linear. Seen languages often perform well regardless of these factors, and some partially seen languages with moderate overlap or fragmentation still achieve relatively high performance. These results indicate that while tokenization characteristics contribute to model behavior, they are not the dominant factors.

\vspace{6pt}
In summary, our tokenization analysis reveals that both over-tokenization and subword overlap influence cross-lingual performance to some extent, particularly for partially seen languages. Nonetheless, these features alone cannot account for the full variation in models' performance. Pre-training familiarity and learned contextual understanding remain the key determinants of performance across different language categories.

\section{Conclusion and Future Work}
This study presents an empirical investigation into the transferability of language models to Indonesian local languages, with a focus on zero-shot and adapter-based approaches for sentiment analysis. By evaluating monolingual (IndoBERT), multilingual (mBERT, XLM-R), and adapter-augmented (MAD-X) models, we demonstrate how language exposure and linguistic similarity influence cross-lingual performance.

Our experiments show that languages seen during pre-training achieve the highest accuracy, partially seen languages perform moderately well, and unseen languages pose the greatest challenge. Adapter-based methods significantly improve performance, especially for seen and partially seen languages, by leveraging even small amounts of unlabeled monolingual data.

Tokenization and vocabulary overlap analyses reveal that while subword fragmentation and shared tokens contribute to performance, they are not the primary drivers of transfer success. Instead, deeper language understanding gained through pre-training or adapter-based adaptation proves more crucial. This insight underscores the importance of using diverse and high-quality language data during model development.

Future work could explore expanding pre-training corpora to include more local languages and adapting tokenizers to better suit low-resource settings. Clustering languages to train shared adapters or stacking general and language-specific adapters might further enhance efficiency. Incorporating knowledge of loanwords and testing few-shot learning methods are also promising directions. Extending the evaluation beyond sentiment analysis to other tasks may also help validate the broader applicability of our findings.

Overall, this study highlights the potential of cross-lingual language model transfer for low-resource languages and emphasizes the importance of both linguistic insight and adaptable computational strategies in building inclusive and effective multilingual NLP systems.

\section*{Acknowledgements}
This work was supported by the Department of Computer Science and Electronics, Universitas Gadjah Mada under the Publication Funding Year 2025.






\input{references}
\end{document}

%% file: tables/main_result.tex
\begin{table*}[htbp]
\caption{Models' Performance (Macro-F1 Score) on Zero-Shot Settings}
\begin{center}
\begin{tabular}{|l|l|c|c|c|c|cccccccc|}
\hline
\textbf{Target} & \textbf{Lang.} & \textbf{Full} & \textbf{IndoBERT} & \textbf{mBERT} & \textbf{XLM-R} & \multicolumn{8}{c|}{\textbf{MAD-X (Target Language Adapter)}} \\
\cline{7-14}
\textbf{Lang.} & \textbf{Category} & \textbf{FT} & & & & \textbf{ind} & \textbf{jav} & \textbf{sun} & \textbf{min} & \textbf{ace} & \textbf{bjn} & \textbf{ban} & \textbf{bug} \\
\hline
jav$^{1,2}$ & Seen & \textit{0.85} & 0.68 & 0.66 & 0.81 & 0.80 & \textbf{0.85} & 0.80 & 0.79 & 0.74 & 0.79 & 0.67 & 0.69 \\
sun$^{1,2}$ & Seen & \textit{0.77} & 0.59 & 0.65 & 0.72 & 0.75 & 0.75 & \underline{\textbf{0.82}} & 0.73 & 0.65 & 0.72 & 0.63 & 0.58 \\
\hline
min$^1$ & Partially-seen & \textit{0.80} & 0.70 & 0.62 & 0.75 & 0.75 & 0.74 & 0.74 & \underline{\textbf{0.83}} & 0.68 & 0.76 & 0.64 & 0.67 \\
ace & Partially-seen & \textit{0.73} & 0.55 & 0.60 & 0.55 & 0.60 & 0.62 & 0.61 & 0.62 & \textbf{0.68} & 0.58 & 0.52 & 0.42 \\
bjn & Partially-seen & \textit{0.80} & 0.79 & 0.65 & 0.75 & 0.74 & 0.75 & 0.74 & 0.77 & 0.69 & \underline{\textbf{0.85}} & 0.60 & 0.63 \\
nij & Partially-seen & \textit{0.73} & 0.54 & 0.59 & 0.61 & 0.58 & 0.60 & 0.58 & \textbf{0.64} & 0.51 & 0.60 & 0.45 & 0.45 \\
mad & Partially-seen & \textit{0.74} & 0.55 & 0.55 & 0.57 & 0.60 & \textbf{0.63} & 0.63 & 0.62 & 0.53 & 0.57 & 0.46 & 0.42 \\
ban & Partially-seen & \textit{0.77} & 0.60 & 0.60 & 0.62 & 0.67 & 0.70 & 0.64 & 0.66 & 0.57 & \textbf{0.74} & 0.46 & 0.46 \\
\hline
bbc & Unseen & \textit{0.70} & 0.39 & \textbf{0.45} & 0.33 & 0.36 & 0.39 & 0.37 & 0.38 & 0.30 & 0.37 & 0.27 & 0.23 \\
bug & Unseen & \textit{0.65} & 0.28 & \textbf{0.43} & 0.28 & 0.31 & 0.32 & 0.31 & 0.29 & 0.23 & 0.29 & 0.24 & 0.18 \\
\hline
\textbf{avg} & - & \textit{0.75} & 0.57 & 0.58 & 0.60 & 0.62 & \textbf{0.64} & 0.63 & 0.63 & 0.56 & 0.62 & 0.52 & 0.47 \\
\hline
\multicolumn{14}{l}{$^{1}$ Language included during pre-training in mBERT. $^{2}$ Language included during pre-training in XLM-R.} \\
\multicolumn{14}{l}{``Full FT'' refers to full fine-tuning baseline of XLM-R.}
\end{tabular}
\label{tab:main_result}
\end{center}
\end{table*}